\documentclass[conference]{IEEEtran}
\usepackage{subcaption}
\usepackage{float}
\usepackage{multirow, multicol}
\usepackage{hhline}
\usepackage{amsmath}
\usepackage{array, makecell}%
\usepackage{bm}
\usepackage{amsbsy}
\usepackage{booktabs}
\usepackage{xspace}
\usepackage{xcolor}
\usepackage{newfloat}
\usepackage[export]{adjustbox}
\usepackage{enumitem}
\usepackage[hidelinks]{hyperref}
\usepackage{url}

\usepackage[nottoc]{tocbibind}
\usepackage{mathtools}

\usepackage[utf8]{inputenc} 
\usepackage{csquotes}
\usepackage{enumitem}
\usepackage{ifthen}
\usepackage{lipsum}
\usepackage{graphicx}
\usepackage{caption}
\usepackage{subcaption}

\usepackage{siunitx}
\def\eg{\textit{e.g.}\@\xspace} 
\def\ie{\textit{i.e.}\@\xspace} 
 
\def\cf{\textit{cf.}\@\xspace}

\def\etal{\textit{et al.}\@\xspace}

\newcommand*{\roundnum}[1]{\num[output-decimal-marker={.},
                             round-mode=places,
                             round-precision=2,
                             group-digits=false]{#1}}
\usepackage{pgfplots}
\usepackage{pgfgantt}
\usepackage{algorithm}
\usepackage[noend]{algpseudocode}
\usepackage{braket}
\IEEEoverridecommandlockouts
\def\BibTeX{{\rm B\kern-.05em{\sc i\kern-.025em b}\kern-.08em
    T\kern-.1667em\lower.7ex\hbox{E}\kern-.125emX}}

\begin{document}

\title{Child Face Recognition at Scale: Synthetic Data Generation and Performance Benchmark\\
{\large Magnus Falkenberg, Anders Bensen Ottsen, Mathias Ibsen, Christian Rathgeb}
\thanks{The authors are affiliated with the Technical University of Denmark and the Biometrics and Security Research Group at Hochschule Darmstadt\\
E-mail: mathias.ibsen@h-da.de \\ 
Magnus Falkenberg and Anders Bensen Ottsen contributed equally}
}
\author{}
\maketitle
\thispagestyle{plain}
\pagestyle{plain}

\begin{abstract}
We address the need for a large-scale database of children's faces by using generative adversarial networks (GANs) and face age progression (FAP) models to synthesize a realistic dataset referred to as HDA-SynChildFaces. To this end, we proposed a processing pipeline that initially utilizes StyleGAN3 to sample adult subjects, which are subsequently progressed to children of varying ages using InterFaceGAN. Intra-subject variations, such as facial expression and pose, are created by further manipulating the subjects in their latent space. Additionally, the presented pipeline allows to evenly distribute the races of subjects, allowing to generate a balanced and fair dataset with respect to race distribution. The created HDA-SynChildFaces consists of 1,652 subjects and a total of 188,832 images, each subject being present at various ages and with many different intra-subject variations. Subsequently, we evaluates the performance of various facial recognition systems on the generated database and compare the results of adults and children at different ages. The study reveals that children consistently perform worse than adults, on all tested systems, and the degradation in performance is proportional to age. Additionally, our study uncovers some biases in the recognition systems, with Asian and Black subjects and females performing worse than White and Latino Hispanic subjects and males.
\end{abstract}

\vspace{10pt}

\section{Introduction}
The use of facial recognition systems in various domains such as surveillance, airports, and personal devices, has been well established. These systems have proven to be highly effective and accurate in verifying the identity of subjects~\cite{facregsurvey, deepfaceregsurvey}. However, as facial recognition becomes increasingly integrated into our daily lives, it is crucial to consider the potential for biases and discrimination against certain demographic groups. Previous research has investigated this issue \cite{DemographicBias}, but less attention has been given to the effect of age on the recognition of children's faces. This area is essential, as there are numerous potential applications for face recognition systems for children. For instance, police can use it to find kidnapped or lost children. Another use case is an automated process for analyzing seized child sexual abuse material (CSAM) to recognize victims. In 2019, more than 70 million CSAM videos and images were obtained\footnote{\url{https://www.europarl.europa.eu/RegData/etudes/BRIE/2020/659360/EPRS_BRI(2020)659360_EN.pdf}}. This issue is an increasing problem with 17 million reports of CSAM in 2019, which saw a dramatic increase to 29.3 million reports in 2021\footnote{\url{https://www.europarl.europa.eu/RegData/etudes/BRIE/2022/738224/EPRS_BRI(2022)738224_EN.pdf}}. Due to this immense amount of data it is necessary to have automated systems for identifying children in such material, which requires effective face recognition systems. 

The emergence of deep learning in recent years has shown to be extremely useful in face recognition~\cite{deepfaceregsurvey}. A caveat of these models is that they need a huge amount of training data to achieve state-of-the-art performance. The amount of data needed has become a growing concern due to increased legal and political scrutiny surrounding the privacy issues associated with large datasets of individuals' faces \cite{RiseAndFallOfDatasets, deprecatingDatasets, Exposing.ai}. The current databases used for research in this area are often limited in size, constrained, and focused on specific ages or races and are frequently retracted due to privacy concerns. This issue is further exacerbated when it comes to children due to heightened focus on protecting their rights.

\begin{figure}
    \centering
    \includegraphics[width=\linewidth]{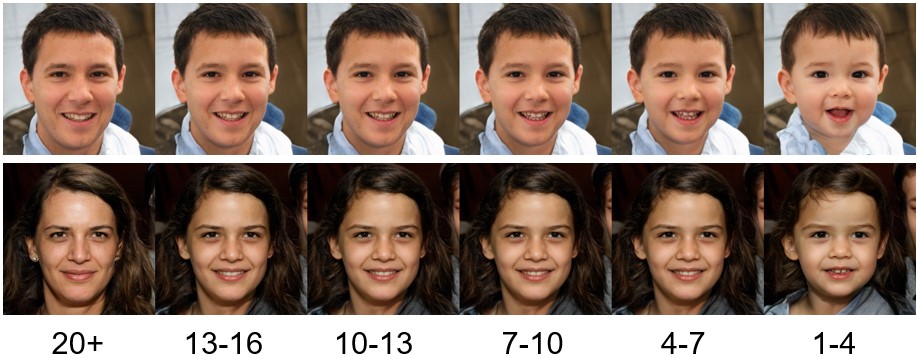}\vspace{-0.1cm}
    \caption{Examples of face images generated by StyleGAN3 \cite{STYLEGAN3} (leftmost) with progressed child faces of varying ages using InterFaceGAN \cite{InterFaceGAN}.}
    \label{fig:intro-example}
\end{figure}

\begin{table*}[htb]
\caption{Different child face datasets for facial recognition systems. The bias column indicates the presence of any potential biases within the respective dataset.  }
    \label{tab:children-datasets}
    \begin{adjustbox}{max width=1\textwidth,center}
    \centering
    \begin{tabular}{|l|c|c|c|c|c|c|c|} \hline
        \textbf{Dataset} & \textbf{Year} & \textbf{\# Identities} & \textbf{\# Images} & \textbf{Age Span}       & \textbf{Bias}  & \textbf{Acquisition Method} & \textbf{Availability}   \\ \hline 
        ITWCC \cite{ITWCC}              & 2015 & 304 & 1,705 &   0.5-32   & $^{a}$ & Scraped           & Public$^{b}$ \\ \hline 
        NITL \cite{NITL}                & 2016 & 314  & 3,144  & 0-4      & Race & Controlled    & Private \\ \hline
        CLF \cite{CLF}                  & 2017 & 919 & 3,682 & 2-8      & Race  & Controlled      & Private \\ \hline
        YFA \cite{YFA}                  & 2022 & 231 & 2,293 & 3-14      & Race  & Controlled        & Public$^b$ \\ \hline
        ICD \cite{childgan}              & 2022 & 16,969 & 35,484 &  2-19  & Race   & Controlled & Private \\ \hline
        YLFW \cite{YLFW}                & 2023 & 3,069 & 9,810 & 0-$\sim$18 & -                    & Scraped    & Not released \\ \hline 
    \end{tabular}
    \end{adjustbox}\vspace{0.1cm}
    \footnotesize{\\$^a$ The authors do not describe the demographic distribution of the dataset.}
    \footnotesize{\\ $^b$ The datasets seem to not be publicly available anymore.}
\end{table*}

To address the aforementioned issues, we present a novel pipeline for creating a synthetic face database containing the same subjects both at adult age, but also at different child ages, see figure~\ref{fig:intro-example}. To do so, state-of-the-art generative adversarial networks (GANs) and face age progression (FAP) models are combined enabling the generation of the first large-scale synthetic child face image database referred to as HDA-SynChildFaces. Two open-source and one commercial face recognition system are evaluated on this database. It is found that the recognition performance of all tested systems decreases with age groups. Evaluations on further demographic subgroups, \ie gender and race, additionally reveal certain biases in the tested face recognition systems. The generated HDA-SynChildFaces dataset, which can be used to train or evaluate face recognition systems, will be made available to the public domain to facilitate reproducible research~\cite{HDASynChildFaces}. Additionally, the source code for the data generation pipeline will also be made available: \url{https://github.com/dasec/HDA-SynChildFaces-AgeTransformation}.%

The rest of this work is organised as follows: section~\ref{sec:related} briefly discusses related works on face recognition for children and face age progression. The database generation process is described in detail in section~\ref{sec:db}. Experiments are presented in section~\ref{sec:experiments} and a discussion is provided in section~\ref{sec:discussion}. Finally, conclusions are drawn in section~\ref{sec:conclusion}.

\section{Related Work}\label{sec:related}

\subsection{Child Face Recognition}

Multiple efforts have been made to create datasets of children at different ages to evaluate or train facial recognition systems. However, many datasets used in research are not publicly available due to ethical and privacy concerns. These efforts are largely divided into two categories: controlled datasets obtained through controlled settings, such as \cite{NITL}, \cite{CLF}, \cite{YFA}, and \cite{childgan}, or web-scraped datasets, such as \cite{ITWCC}, \cite{YLFW}. An overview of the relevant datasets and their statistics is presented in table~\ref{tab:children-datasets}.

In general, the controlled datasets are obtained in environments where the researchers control different factors such as pose, facial expression, illumination, and the age gap between sessions. This makes it easier to isolate the dataset to only focus on age differences. However, one limitation of these datasets is the potential for race and demographic bias in the sample population. The web-scraped datasets are often less constrained and have more variation in the images. This makes it more difficult to distinguish between the effects of age and other factors on the performance of facial recognition systems. Most of the datasets are used for longitudinal studies of the performance of facial recognition systems.
The NITL \cite{NITL} is a longitudinal dataset focusing on children aged 0-4. The data were collected at a free paediatrics clinic in Dayalbagh, India. The images were collected over four sessions between March 2015 and March 2016. Their experiment compares facial recognition systems' accuracy on images from the same sessions with images from different sessions. They found that the verification accuracy when children aged six months decreased with 50\% compared to the verification of images taken in the same session. The difference was even more significant when only looking at children aged 1-2 in the first session. Here the accuracy was decreased by 82\%.

In the three longitudinal studies \cite{CLF, YFA, childgan}, the datasets were collected in cooperation with schools. These datasets are not public available due to privacy reason regarding the subjects.
In \cite{CLF}, the CLF dataset consists of facial images of children aged 2-18 with constrained images taken of the same subject over time (avg. 4.2 years). The YFA \cite{YFA} dataset contains images captured over time at a local elementary and middle school of voluntary children. In the dataset the target is to investigate how changes in ages influence facial recognition systems, and thus the images captured are limited with regards to change in pose, illumination and expression. The images are taken over multiple session with a maximum total age gap of 3 years. 
The ICD dataset, used in ChildGAN \cite{childgan}, contains subjects with multiple images taken over time as well as subjects with only a single image. The facial images are divided into five different sets based on the following age groups: 2-5, 6-8, 9-11, 12-14, and 15-19. All of the images were collected in India. 
The In-the-Wild Child Celebrity (ITWCC) \cite{ITWCC} dataset is a recent longitudinal children database scraped from the internet. The dataset consist of different celebrities. As the images are scraped from the internet, they are unconstrained, this makes it difficult to isolate the age as a parameter when testing facial recognition systems. They present results showing that facial recognition systems has issues verifying identities of non-adult aging subjects. 
Another, and very recent, dataset scraped from the internet is the YLFW \cite{YLFW} dataset. In this, the authors have a method of scraping identities on the web using a specific set of keywords. A set of images are downloaded for each of these keyword sets. This set of images are then filtered by using hierarchical clustering. They then balance the dataset regarding the four races: Caucasian, Asian, African, and Indian. A manual procedure follows this process to verify the match pairs. In the performance evaluation of facial recognition systems for children, they find that the systems are significantly worse for children, as previously studies also have shown. However, they also show that training facial recognition systems on their dataset can reduce this difference.

In \cite{faceRecBiasOSTI}, a subsets of the ITWCC \cite{ITWCC} and LFW \cite{LFW} datasets are used to compare the performance of facial recognition systems performance on adults and children. The authors compare 8 different facial recognition systems and find that all eight facial recognition systems were biased, performed significantly worse on children. 

\subsection{Face Age Progression}
GAN-based architectures have not only proven their worth in generating synthetic images but also for performing face age progression (FAP). Grimmer \etal~\cite{9446061} have recently provided a comprehensive survey on deep face age progression, noting that GANs indeed produce \textit{remarkable face ageing results}, \cf figure \ref{fig:intro-example}. Many of the covered FAP models in this section are based on GANs in some way. 

In InterFaceGAN~\cite{InterFaceGAN}, the authors do not directly train a new GAN to do FAP but instead investigate the latent space learned by the original StyleGAN trained on the FFHQ dataset. The researchers train a linear model in which a boundary is learned to, \eg change the age or gender of a generated image directly in the latent space. In~\cite{TTTC}, Alaluf \etal retrain InterFaceGAN on the StyleGAN3 latent space thus taking advantage of the improved architecture for generating faces.

In~\cite{LIFESPAN}, the authors handle FAP by proposing a new GAN architecture trained with labelled age groups (\eg 0-2 or 50-69), enabling FAP by giving an input image and specifying the wanted age group. He \etal \cite{DLFS} note that many of the GAN-based FAP approaches end up with an \textit{entangled} latent space in which they then manipulate the age. In their work, they instead propose a model where they disentangle key characteristics while modifying the age, also in different age groups. In~\cite{childgan}, the authors also take a GAN-based approach to FAP learned with different age groups. AgeGAN, an architecture proposed in ~\cite{AgeGAN++}, uses a dual condition GAN architecture, where one generator converts input faces to other ages based on an age group condition, and the dual conditional GAN learns to invert the task. 

In~\cite{SAM}, Alaluf \etal propose an architecture where age is approached as a regression task, rather than discrete age groups. Their model learns a non-linear path to disentangle the age progression from other attributes. Li \etal~\cite{CFASR} also focus on continuous ageing and use an age estimator as part of a GAN-generator in a novel architecture. 

Authors from Disney Research in~\cite{disney_fap} propose a FAP model that does not use a GAN but instead a U-Net, translating in an image-to-image manner together with a provided age, and note promising results. A caveat of their model is that it is only possible to progress down to the age of 20, due to the training data used.

Many of the models proposed in the scientific literature, \eg \cite{LIFESPAN, DLFS, SAM, disney_fap}, are all end-to-end solutions, meaning that they take an image and a wanted age (or age-group) as input and then outputs an image. In~\cite{InterFaceGAN}, and thus also in~\cite{TTTC}, an image is manipulated directly in the latent space of the StyleGAN variant, which then skips the part of inverting or translating an image into latent space, which otherwise may come at a loss.

\section{Database Generation}\label{sec:db}
The proposed pipeline for creating the desired biometric dataset consists of the following steps that are described in detail in the subsequent subsections: 

\begin{description}
    \item[Sampling] This step handles the generation of synthetic faces, thus creating the initial database. 
    \item[Filtering] The filtering step handles the filtering of the initial database, removing poor quality and unwanted images. 
    \item[Race Balancing] As the generation of the initial faces is random, the distribution of the subjects' races may be skewed. This step evenly distributes races in the database. 
    \item[Age Transformation] Progressing an adult into a child is a key concept in this paper. This step does that by progressing an adult into a child in different age groups. 
    \item[Intra-Subject Transformation] To biometrically benchmark a database, reference images need corresponding probe images, which this step is responsible for creating. 
    \item[Post-Processing] This step will do some automatic cleaning. It ensures that the same seeds are present in all different age groups and tries to remove poorly transformed images. 
\end{description}

\subsection{Sampling and  Filtering}
In this work, StyleGAN3 is used to sample an initial set of face images. A subset of these initially sampled images are then chosen by performing a filtering by first discarding images based on age and afterwards based on sample quality.

The age filtering step is implemented by using the C3AE age estimator \cite{C3AE}. It simply works by estimating the age of the generated subjects and if they are below a pre-defined age they are rejected. 
\begin{figure}[!htbp]
    \centering
    \begin{subfigure}{0.48\linewidth}
    \includegraphics[width=\linewidth]{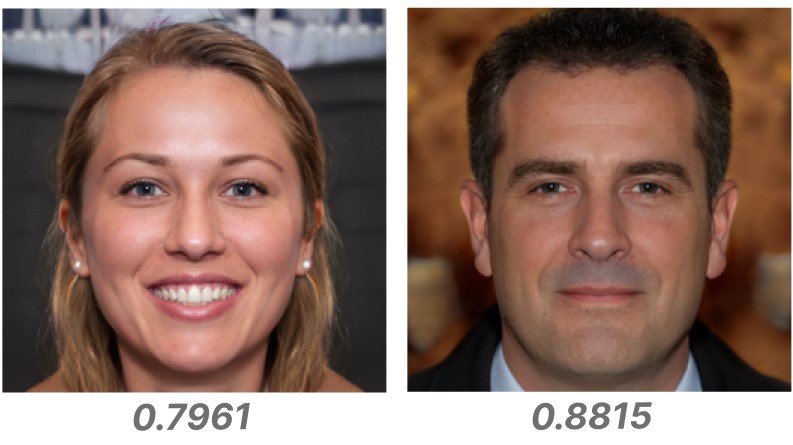}
    \caption{accepted}
    \label{fig:impl:quality:accepted}
    \end{subfigure}
    \begin{subfigure}{0.48\linewidth}
    \includegraphics[width=\linewidth]{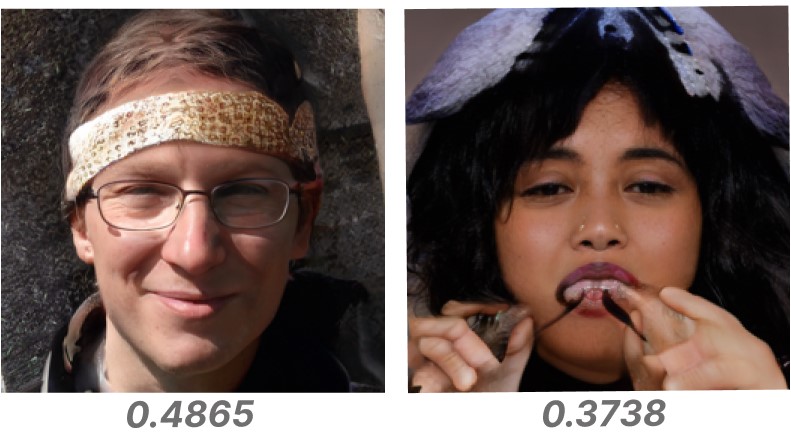}
    \caption{rejected}
    \label{fig:impl:quality:rejected}
    \end{subfigure}\vspace{-0.1cm}
    \caption{Example of accepted and rejected images using SER-FIQ \cite{facequality} quality assessment with a threshold of 0.75.}
    \label{fig:impl:fq_examples}
\end{figure}

\begin{figure}[!htbp]
    \begin{center}
    \scalebox{0.7}{
     \input{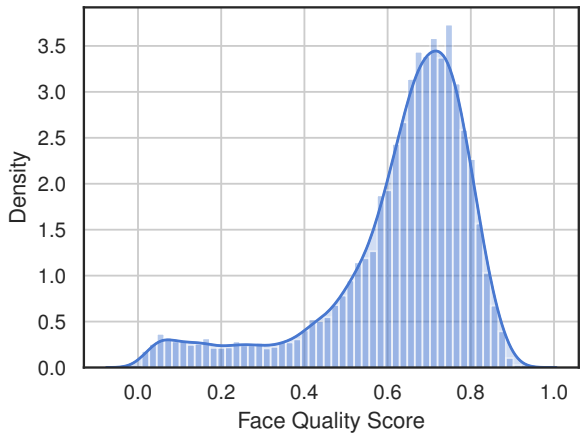}}
    \end{center}\vspace{-0.3cm}
    \caption{Distribution of SER-FIQ~\cite{facequality} quality scores for 10,000 generated face images.}\label{fig:software_dev:fq_dist}
\end{figure}

The quality filtering step is implemented by using the SER-FIQ~\cite{facequality} quality score algorithm, which represents a state-of-the-art algorithm. The quality score extracted is between 0 and 1, where  1 is an image of perfect quality. Figure \ref{fig:impl:fq_examples} shows examples of accepted and rejected images based on the SER-FIQ score.  Figure~\ref{fig:software_dev:fq_dist} shows the distribution of the quality scores for a 10,000 generated subjects without doing any age filtering beforehand. The distribution looks Gaussian-like but with a heavy tail that may be caused by artefacts or very young looking subjects which usually gets a low quality score.
\raggedbottom

\subsection{Latent Transformation}

\begin{figure}[!htbp]
    \centering
    \includegraphics[width=0.8\linewidth]{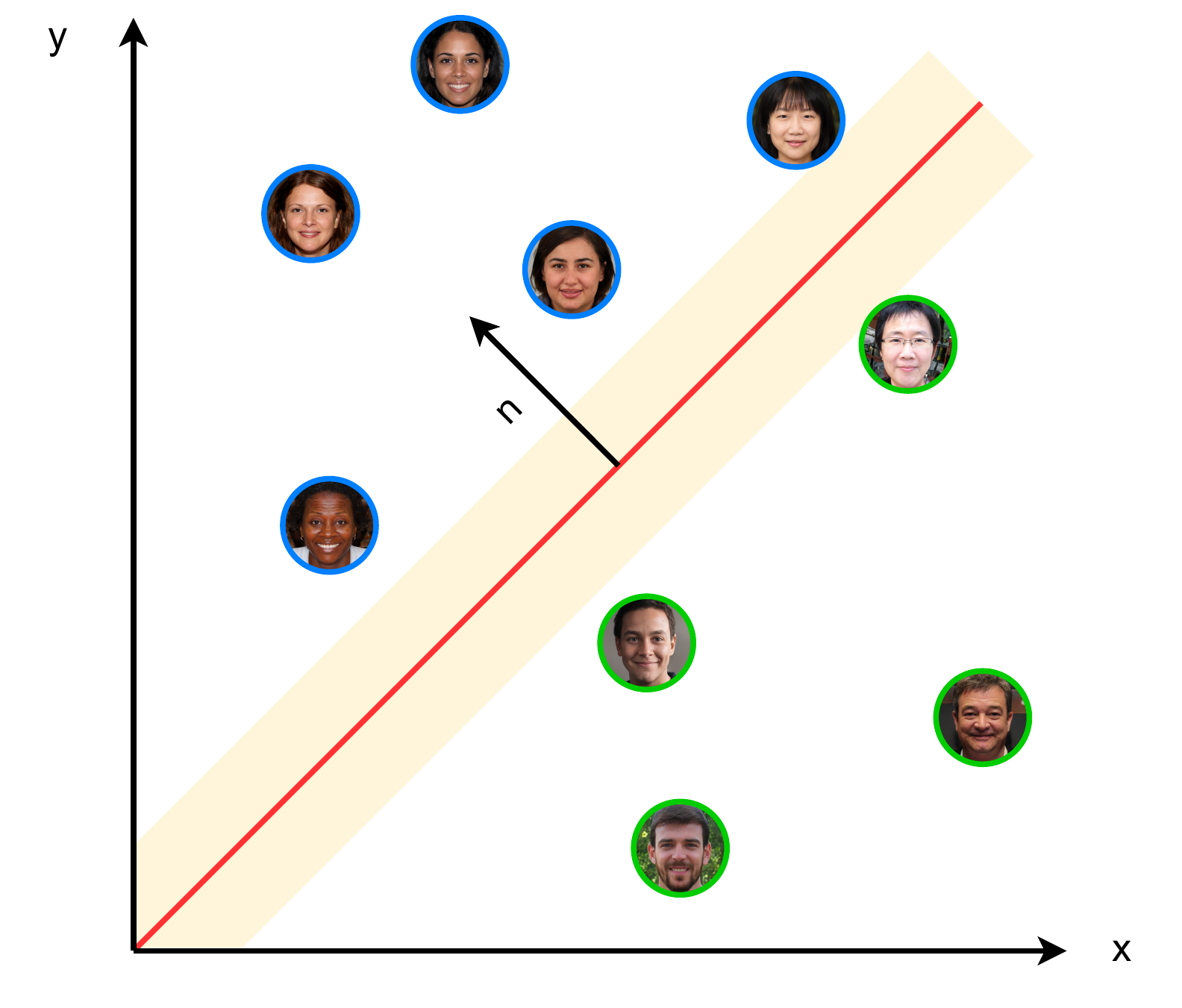}
    \caption{Hyperplane between two categories in a simplified 2D space. The red line is the hyperplane as found by a linear SVM. The images with a blue border are categorized as women, and those with a green border are categorized as men. The vector $n$ is the normal vector to the hyperplane.}
    \label{fig:interfacegan-2d}
\end{figure}

\begin{figure}
    \centering
    \includegraphics[width=\linewidth]{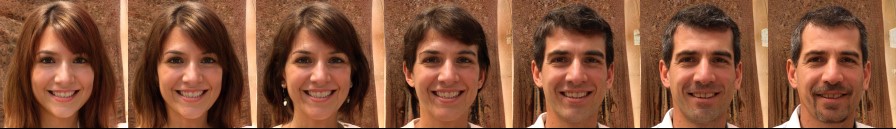}
    \caption{An example of a subject manipulated with a gender boundary, moving the subject in both a positive and negative direction}
    \label{fig:igan-gender}
\end{figure}

The method used for estimating hyperplane boundaries for certain attributes is done as explained in the original InterFaceGAN paper~\cite{InterFaceGAN}, but by using the latent space of StyleGAN3 instead of StyleGAN. The separation boundary between different categories of an attribute is found by the use of a linear support vector machine (SVM) to identify a hyperplane that separates the two categories. For example, in the case of the gender attribute, the SVM could be trained to distinguish between Male and Female, see figure \ref{fig:interfacegan-2d}.
This normal vector $n$ can then be used to modify the latent code of an image by adding it to the latent code. This can be described as:
\begin{equation}
    w_{edit} = w + \alpha \cdot n
\end{equation}
where $w$ is the latent code of the image, $\alpha$ is a parameter choosing the degree of the edit and $w_{edit}$ is the resulting latent code after the manipulation. figure \ref{fig:igan-gender} shows how this boundary modifies a subject, in the example the same subject is manipulated with different $\alpha$ values.

SVM are trained on a large amount of images (500,000) generated with StyleGAN3. Each image must be classified using a pre-trained classifier for the specific attribute. To filter out bad classifications only the top 10\% and bottom 10\% are used for the training. This was done for all of the attributes in table \ref{tab:impl:all-boundaries} marked with \textit{This work}. 

\begin{table}[!htbp]
 \caption{Boundaries used in the implementation of the proposed pipeline.}
    \label{tab:impl:all-boundaries}
    \begin{adjustbox}{max width=1\linewidth,center}
    \begin{tabular}{|l|l|l|l|}
    \hline
         \textbf{Category} &\textbf{Attribute} & \textbf{Classifier} & \textbf{From}  \\ \hline 
          \multirow{2}{*}{Age} & Age & C3AE \cite{C3AE} & This work \\ \cline{2-4} 
         & Age & SAM age estimator \cite{SAM} & \cite{TTTC} \\ \hline
         \multirow{2}{*}{Pose}&Yaw & Hopenet \cite{HOPENET} & This work\\ \cline{2-4} 
        &  Pitch &  Hopenet \cite{HOPENET} & This work \\ \hline
       \multirow{2}{*}{Expression} & Happy & Anycost GAN \cite{lin2021anycost} & \cite{TTTC} \\ \cline{2-4} 
         &Sad & Deepface \cite{deepface1} & This work\\ \hline
         \multirow{6}{*}{Race}&White & Deepface \cite{deepface1} & This work \\ \cline{2-4} 
        &Latino Hispanic & Deepface \cite{deepface1} & This work\\ \cline{2-4} 
        &Indian & Deepface \cite{deepface1} & This work\\ \cline{2-4} 
        &Middle Eastern & Deepface \cite{deepface1} & This work \\ \cline{2-4} 
        &Asian & Deepface \cite{deepface1} & This work \\ \cline{2-4} 
        &Black & Deepface \cite{deepface1} & This work \\ \hline
        Illumination &Illumination &  DPR~\cite{IlluminationZhou} &  This work \\ \hline
        Gender & Male & Anycost GAN \cite{lin2021anycost} & \cite{TTTC} \\ \hline 
         
    \end{tabular}
    \end{adjustbox}
\end{table}

As mentioned in \cite{InterFaceGAN} the manipulation of a specific attribute can  result in unintended changes to other attributes. This is due to the entanglement in the latent space and the correlation of the attributes in the images used for training the SVM. To minimize these unwanted side effects, a new conditional boundary can be calculated by projecting the boundary of the desired attribute onto the boundary of another attribute. This process can be formalized mathematically as follows:
\begin{equation}
    n_{cond} = n_1 - (n_1^T \cdot n_2) \cdot n_2
\end{equation}
Where $n_1$ is the boundary for the desired attribute (\eg smile), $n_2$ is the boundary for the unintended attribute (\eg glasses) and $n_{cond}$ is the conditional boundary. This new conditional boundary can then be used to edit  the desired attribute. Furthermore, the pipeline uses the $w$ latent vector, which is a single dimension of the $w+$ latent. This is due to less entanglement, than when using the $z$ latent, as mentioned in the original InterFaceGAN paper~\cite{InterFaceGAN}.  
 
The need for neutralizing images with respect to certain attributes, such as a pose, arises during image sampling in order to ensure the quality of the generated images. Here, we follow the process proposed in \cite{intra_subject_latent_space}.  If one wanted to neutralize an image with respect to yaw using a trained boundary denoted as $n_{yaw}$, it can be described mathematically as:
\begin{equation}
w_{neutral} = w - (w^{T}n_{yaw}) \cdot n_{yaw}
\end{equation}
where $w$ is the initial latent code for the image and $w_{neutral}$ is the latent vector for the neutralized image. One could then use $w_{neutral}$ to generate the neutralized image. This concept of neutralization is used several times throughout the pipeline, and can be done with any of the boundaries seen in table~\ref{tab:impl:all-boundaries}.

\subsection{Balancing Races}

\begin{figure}[htb]
    \makebox[\linewidth][c]{
    \includegraphics[width=1\linewidth]{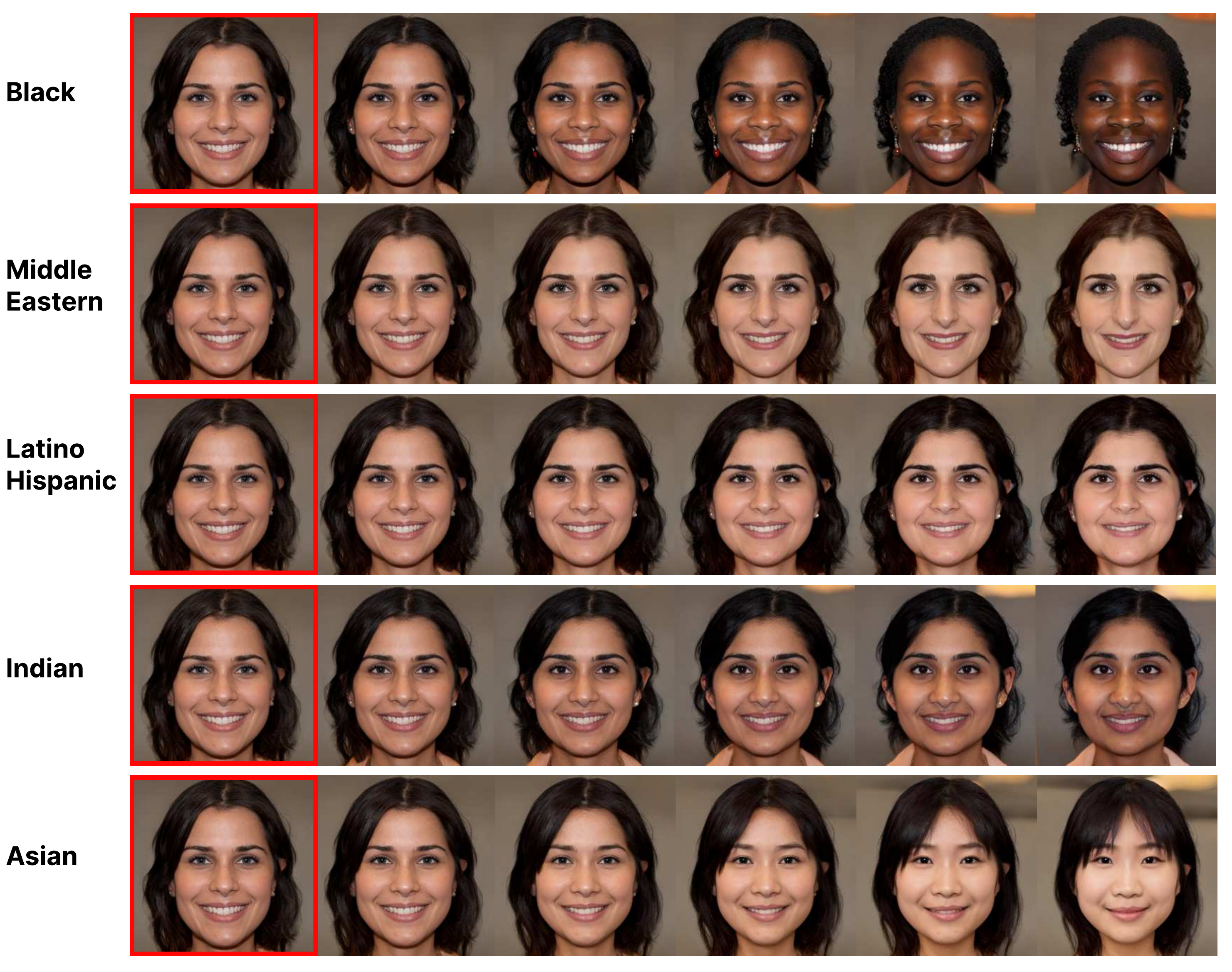}
    }
    \caption{An example of moving a subject (leftmost) along each of the 5 race boundaries.}
    \label{fig:impl:races_examples}
\end{figure}

In contrast to real existing child face databases, we aim at creating a database that is equally distributed with respect to race. To do so the trained race boundaries as seen in table~\ref{tab:impl:all-boundaries} can be used to change the race of individual subjects. Figure~\ref{fig:impl:races_examples} shows examples of using the individual race boundaries on the same subject.

Firstly, a database of images and latent vectors is sampled, where the race of each of the subjects is initially classified. Subsequently, a random subject of the most represented race is changed into the least represented race. This step is repeated until the races are uniformly distributed. 

\begin{figure}
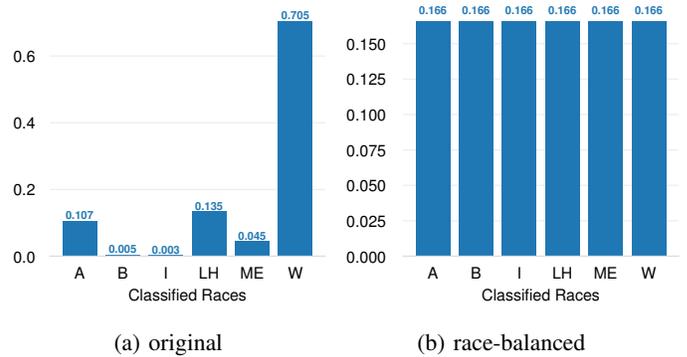

     \centering
     \begin{subfigure}[b]{0.5\linewidth}
        \centering
           \scalebox{0.6}{\input{figures/dist_before.pgf}}
         \caption{original}
         \label{fig:race_algo_prior}
     \end{subfigure}%
     \begin{subfigure}[b]{0.5\linewidth}
        \centering
        \scalebox{0.6}{\input{figures/dist_after.pgf}}
        \caption{race-balanced}
        \label{fig:race_algo_post}
     \end{subfigure}
        \caption{Effect of race balancing on a dataset of 3510 subjects of non-uniformly distributed races. \textit{A: Asian}, \textit{B: Black}, \textit{I: Indians}, \textit{LH: Latino-Hispanic}, \textit{ME: Middle Easteren}, \textit{and} \textit{W:White}.}
        \label{fig:race_algo}
\end{figure}

An example of the distribution of the races before and after balancing races can be seen in figure~\ref{fig:race_algo}. From the figure it can be seen that, initially, 70\% of the subjects sampled are classified as white, where only 0.5\% are classified as black. It should be noted that a caveat of this approach is that it is largely dependant on the race classifier. That is, human inspection of the subjects races may not always agree with what the classifier and algorithm outputs. 

\subsection{Age Transformation}

\begin{figure}
    \centering
    \includegraphics[width=1\linewidth]{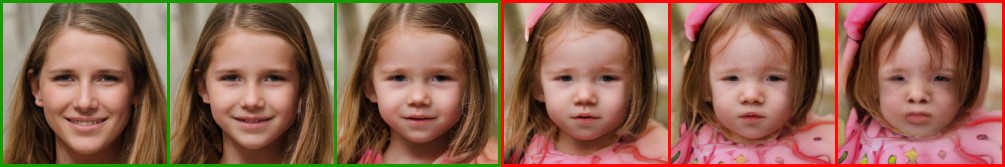}
    \caption{Example of a subject moved too far in the age direction (green boxed images still look natural, where the red boxed images looks progressively more unnatural).}
    \label{fig:impl:subject_too_far}
\end{figure}

The latent transformations previously described is also used to transform the age of a subject. However, one problem of the latent transformations is that sometimes a subject is transformed poorly because the subject is moved too far in a direction in the latent space. This can, for instance, happen if the age classifier inaccurately predicts the age of a subject. An example of a subject being moved too far can be seen in figure~\ref{fig:impl:subject_too_far}. The first 3 images, surrounded by green boxes, are realistically looking and looks like the same person in progressively younger versions. In the last 3 images, surrounded by red boxes, it can be seen that by moving too far along the age direction the subject starts to look less human and unrealistic, \ie poorly transformed. 

\begin{figure}[t!]
    \centering
    \includegraphics[width=0.8\linewidth]{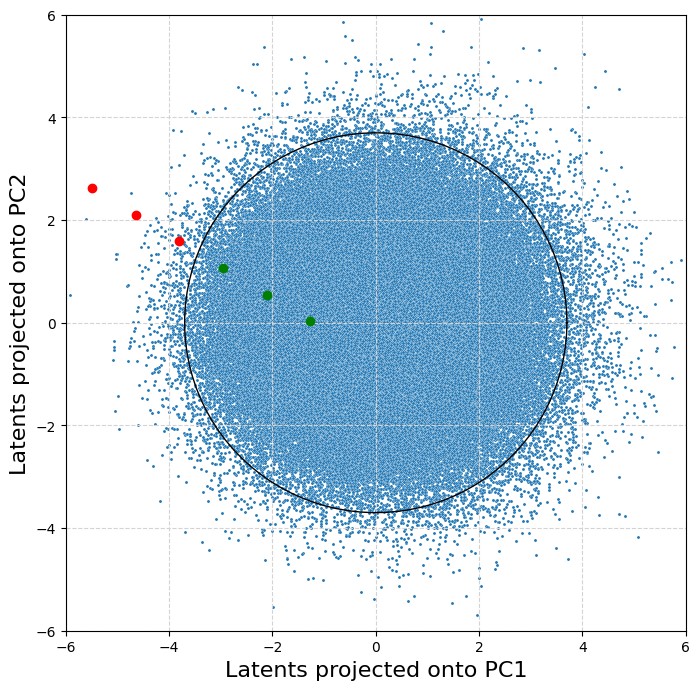}
    \caption{300k StyleGAN3 latent vectors projected onto their first and second principal components, where each blue dot corresponds to a subject. The red and green dots showcases the effect of transforming a single subject along the age direction, corresponding to the progression seen in figure~\ref{fig:impl:subject_too_far}, where the red dots are going out of the distribution.}
    \label{fig:impl:pca_200k}
\end{figure}

A way to automatically detect such undesired effects is to do a principal component analysis (PCA) and use it for outlier detection. We generated a large amount of latent vectors (300,000) to fit the PCA and find the principle components. The idea is that the two most important principal components form a distribution and that a transformed image is an anomaly if it is too far away from the center. If the image is categorized as an anomaly, then it should be removed as it is likely to be one of poor transformation. A visualization of this concept can be seen in figure~\ref{fig:impl:pca_200k}.

\begin{figure}
    \centering
    \includegraphics[width=\linewidth]{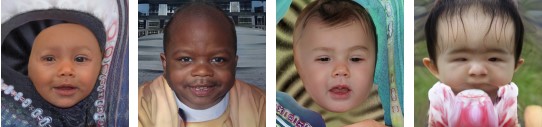}
    \caption{Examples of rejected images in the automatic post-processing step of the pipeline.}
    \label{fig:pca:outlier-dectection}
\end{figure}

This approach is able to automatically detect the majority of poorly transformed subjects. Figure \ref{fig:pca:outlier-dectection} shows more examples of removed subject from the database where the images are categorized as anomalies.

\subsection{Intra-Subject Transformations}
\begin{figure}
    \centering
    \includegraphics[width=0.9\linewidth]{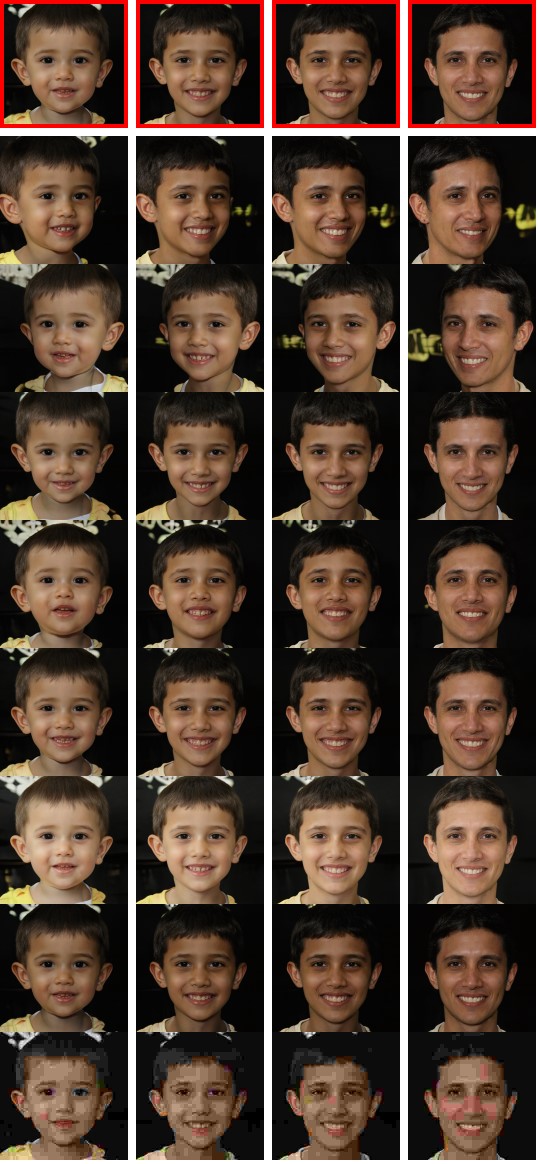}
    \caption{Example of a subject in the different age groups. The ages from left to right: \textit{1-4},  \textit{7-10}, \textit{13-16} and \textit{20+}. The variations from the top and downwards, where the reference is the red boxed ones: \textit{left yaw}, \textit{right yaw}, \textit{pitch down}, \textit{pitch up}, \textit{smile}, \textit{high illumination}, \textit{low illumination} \& \textit{compression}.}
    \label{fig:dataset:subject-example}
\end{figure}

The following subject- and environment-related properties are further modified to simulate intra-class variations: pose, expression, and illumination.These are implemented by manipulating the latent vectors using the linear boundaries trained as seen in table~\ref{tab:impl:all-boundaries}. Figure \ref{fig:dataset:subject-example} depicts all variations across different age groups for an example subject.

For changing the pose of the subject two boundaries were trained for yaw and pitch. The boundaries were trained using the Hopenet pose estimator \cite{HOPENET}. 
By default the pipeline will generate four variations for each axis of the subjects. The amount of illumination in an image is also a boundary trained by using the light classifier from the DPR model by Zhou \etal~\cite{IlluminationZhou}. To change the facial expression of a subject two boundaries were used, one for making a subject smile and one for making a subject look sad. To compress an facial image lossy compressed versions of each subject is generated by saving the image in the JPEG format with different qualities. It should be noted that the original reference image are saved in the lossless PNG format.

\subsection{HDA-SynChildFaces}

\newcommand{\numberOfTotalImages}{188,328 }
\newcommand{\numberOfTotalSubjects}{1,652 }
\newcommand{\numberOfTotalVariations}{18 }
The HDA-SynChildFaces database consists of \numberOfTotalSubjects different subjects which have been processed in the whole pipeline explained before. A short overview of the used parameters can be seen in table~\ref{tab:overview_dataset}.

\begin{table}[!htbp]
    \caption{Age groups, races and intra-subject variations provided as input to the pipeline to produce the dataset.}
    \label{tab:overview_dataset}
    \centering
    \begin{tabular}{|l|l|}
        \hline 
        \textbf{Age Groups} &   20+, 16-13, 13-10, 10-7, 7-4, 4-1 \\\hline
        \textbf{Races} &  \begin{tabular}{@{}l@{}}Asian, Black, Latino Hispanic, Middle Eastern, \\Indian, White \end{tabular}  \\\hline
        \textbf{Variations} &  Yaw, Pitch, Smile, Sad, Illumination, Compress \\\hline 
    \end{tabular}
\end{table}
Here the original \numberOfTotalSubjects subjects correspond to being of age 20 and above. Each of these subjects have been progressed down into the five different age groups seen above, resulting in 6 datasets (1 of adults and 5 of children). Each of these images across the six datasets have 18 corresponding intra-class variations. This sums up to a total of: 1,652 $\times$ 6 $\times$ (18+1) = 188,832 images.

\subsubsection{Gender Subset}\label{section:setup:gender}
As a part of the pipeline each synthetic subject has also been classified as either being male (M) or female (F). This split is done to test if the performance on the face recognition systems between gender, and if it varies across different age groups. The number of images in each group can be seen in table \ref{tab:gender-subset-count}.
\begin{table}[!htbp]
 \caption{Female (F) and male (F) subjects and total images in the database.}
    \label{tab:gender-subset-count}
    \centering
    \begin{tabular}{|l|l|l|}
        \hline
        \textbf{Gender}  & \textbf{Subjects}  & \textbf{Total images}  \\\hline
        Female  &  667         &     76,038          \\\hline
        Male    &   985        &     112,290          \\\hline 
    \end{tabular}
\end{table}
It can be seen that 40.3\% of the subjects are female and thus 59.7\% are male, which is a bit skewed. This skewness happens as a part of the filtering process where the quality filterer is slightly biased against women. 

\subsubsection{Race Subset}
The race of the different subjects are also saved after equally distributing them. This allows for dividing the dataset into race-specific subsets, to see if face recognition systems are biased against some races, and if that changes appear across age groups. The number of images and different subjects for each subset can be seen in table \ref{tab:race-subset-count}.
\begin{table}[!htbp]
\caption{The amount of subjects of the different races in the database.}
    \label{tab:race-subset-count}
    \centering
    \begin{tabular}{|l|l|l|}
         \hline
        \textbf{Race}  & \textbf{Subjects}  & \textbf{Total images}  \\ \hline 
        Asian (A)   & 248          &   28,272            \\ \hline
        Black (B)    &  283        &   32,262            \\ \hline
        Indian (I)    &  276         &   31,464            \\ \hline
        Latino Hispanic (LH)    &   281        &  32,034             \\ \hline
        Middle Eastern (ME)    &   278        &  31,692             \\ \hline
        White (W)  &       286    &   32,604            \\ \hline 
    \end{tabular}
\end{table}
Although the races are equally distributed after the race distribution step, this may become a bit unbalanced due to the post-processing step. For instance, as seen in the table, there are fewer Asians left at the end of the pipeline than of other races. 

\section{Experiments}\label{sec:experiments}

\subsection{Experimental Setup}
The HDA-SynChildFaces database will be evaluated with multiple facial recognition systems to determine the performance difference of facial recognition systems on children. The impact that race and gender has is also evaluated by said systems to investigate whether age may impact these factors as well. The facial recognition systems under investigation are ArcFace \cite{ARCFACE}\footnote{\url{https://github.com/deepinsight/insightface}} and MagFace \cite{magface}\footnote{\url{https://github.com/IrvingMeng/MagFace}}, and a commercial off-the-shelf (COTS) solution. Before doing the facial recognition with both state-of-the-art open-source systems, face detection and alignment was done with RetinaFace~\cite{retinaface}, a state-of-the-art face detection system. %

To evaluate the different recognition systems, biometric measures and metrics from the ISO/IEC 19795-1~\cite{iso_biometric} standard will be used. For the open-source systems the mated (genuine) scores and non-mated (impostor) scores are calculated for each of the datasets by using the cosine similarity measure as seen in equation \ref{eq:consine-sim}.
\begin{equation}
    \label{eq:consine-sim}
    cosine \ similarity = \frac{\sum_{i=1}^{n}A_i B_i}{\sqrt{\sum_{i=1}^{n}A_i^2}\sqrt{\sum_{i=1}^{n}B_i^2}}
\end{equation}
Here $A_i$ and $B_i$ refers to specific feature vectors extracted by a face recognition system. The COTS system uses its own proprietary similarity score. The mated comparisons are done by calculating the similarity score between each of the images with each of its corresponding variations. The non-mated comparisons are done by calculating the similarity score between a image and a random image from all other individuals. The results will be evaluated by using the following metrics:

\begin{description}
\item[FMR/FNMR] Following ISO/IEC 19795-1~\cite{iso_biometric}, False Match Rate (FMR) and False None Match Rate (FNMR) are technical terms used to describe the performance of biometric systems. Specifically, FMR represents the percentage of non-mated comparisons that are incorrectly confirmed as matches at a specific threshold, while FNMR represents the percentage of mated comparisons that are incorrectly rejected as non-mated. In this experiment, the focus will be on evaluating the FNMR values under three distinct conditions, corresponding to FMR values of 0.01, 0.1, and 1 percent.
\item[DET-curves] The Detection Error Trade-off (DET) curve is a plot to visualize the trade-off between the False None Match Rate (FNMR) and the False Match Rate (FMR).
\item[EER] The Equal Error Rate (EER) are the rate where the value of FMR and FNMR are equal. 
\item[Distribution Statistics] The following common distribution statistics will be calculated to characterize the distribution of the mated and non-mated comparisons: \textit{mean $\mu$} and \textit{standard deviation $\sigma$}.
\item[Decidability Index] The decidability index, denoted as $d'$, can be interpreted as a value that describes the amount of separation between two distributions. It will be calculated for the distributions of the mated and non-mated comparisons, where a larger value means a better separation between the two. It is calculated using the following formula~\cite{dprime}: 
\begin{equation}
    d' = \frac{|\mu_{m} - \mu_{nm}|}{\sqrt{\frac{1}{2}(\sigma_{m}^2 + \sigma_{nm}^2)}}
\end{equation}
where $\mu_{m}$ and $\sigma_{m}$ are the mean and standard deviation for the mated comparisons and $\mu_{nm}$ and $\sigma_{nm}$ are the mean and standard deviation for the non-mated comparisons. 
\end{description}

\begin{figure*}[t]
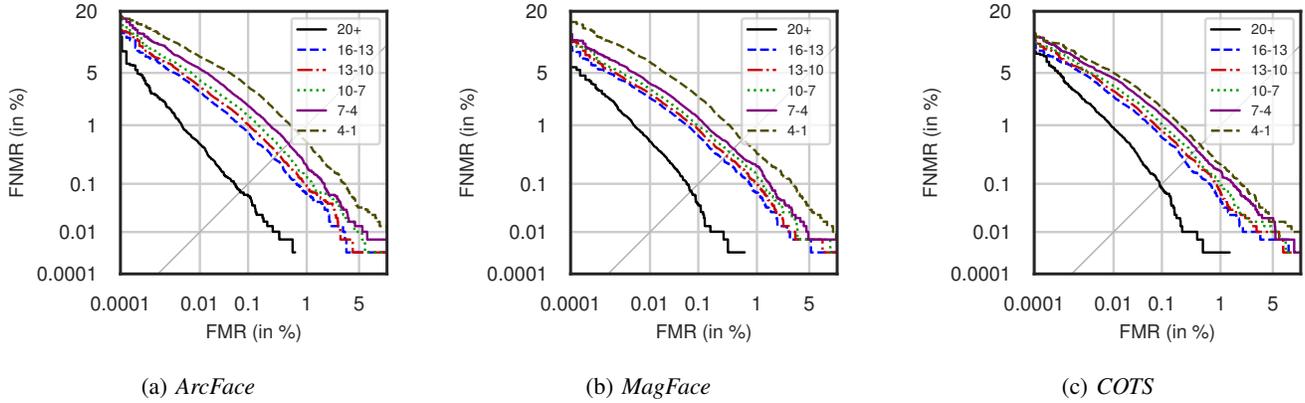

    \centering
    \begin{subfigure}{.33\textwidth}
        \centering
        \scalebox{0.9}{\input{figures/All_DET_curve_arcface.pgf}}
        \caption{\textit{ArcFace}}
        \label{fig:det-full-dataset-arcface}
    \end{subfigure}%
    \begin{subfigure}{.33\textwidth}
     \centering
         \scalebox{0.9}{\input{figures/All_DET_curve_magface.pgf}}
        \caption{\textit{MagFace} }
        \label{fig:det-full-dataset-magface}
    \end{subfigure}
    \begin{subfigure}{.33\textwidth}
        \centering
        \scalebox{0.9} {
        \input{figures/All_DET_curve_cognitec.pgf}}
        \caption{\textit{COTS}}
        \label{fig:det-full-dataset-cots}
    \end{subfigure}
    \caption{DET-curves of the full dataset when using \textit{ArcFace}, \textit{MagFace} \& \textit{COTS}.}
    \label{fig:det-full-dataset-all}
\end{figure*}

\subsection{Results}

\subsubsection{Children vs Adults}

Experimental results across all age groups for the tested face recognition systems are summarised in table~\ref{tab:full-dataset-det-stats}. The corresponding DET curves are plotted in figure~\ref{fig:det-full-dataset-all}. When focusing on one system at a time, it can be seen how the mated part of the table's statistics is very similar across the age groups. The mean and standard deviation are all extremely close. It can be noted how the mated mean and standard deviation are pretty similar for MagFace and ArcFace while COTS has quite a larger mean and smaller standard deviation. When looking at the non-mated part of the table, it can be seen how the mean of the distributions grows steadily the younger the age group, which happens for all three face recognition systems. The same is true for the standard deviation where there is a considerable increase when comparing the age groups 20+ and 16-13. For ArcFace and MagFace, a significant increase between the groups 7-4 and 4-1 can also be noticed, but not for COTS.  Another interesting statistic can be seen when looking at $d'$, where a higher $d'$ means that the system can better distinguish between the mated and non-mated distributions. %
An evident tendency across all systems is that the $d'$ scores decrease for the younger age groups.

\begin{table}[!htbp]
    \caption{Different biometric performance metrics with \textit{ArcFace}, \textit{MagFace} and \textit{COTS} as face recognition systems from the full dataset.}
    \label{tab:full-dataset-det-stats}
    \begin{adjustbox}{max width=\columnwidth,center}
 \begin{tabular}{|c|c|c|c|c|c|c|c|c|c|c|c}
        \hline
        \textbf{Age}  & \multicolumn{2}{c|}{\textbf{Mated}}  & \multicolumn{2}{c|}{\textbf{Non-mated}}  & \textit{EER} & \multirow{2}{*}{\textit{d'}} & \multicolumn{3}{c|}{\textit{FNMR at FMR} (\%)}    \\ %
        \textbf{groups} &  $\mu$ & $\sigma$ &  $\mu$ & $\sigma$ & (\%)  & & \textit{0.01} & \textit{0.1}  & \textit{1} \\ \hline
    \multicolumn{10}{|c|}{\textit{ArcFace}} \\  \hline 
    20+ & \roundnum{0.8848} & \roundnum{0.0955} & \roundnum{0.0602} & \roundnum{0.0889} & \roundnum{0.0711} & \roundnum{8.9336} & \roundnum{0.4978} & \roundnum{0.0610} & \roundnum{0.0000} \\ \hline
16-13 & \roundnum{0.8844} & \roundnum{0.0953} & \roundnum{0.0957} & \roundnum{0.1152} & \roundnum{0.2937} & \roundnum{7.4629} & \roundnum{2.8497} & \roundnum{0.7496} & \roundnum{0.0642} \\ \hline
13-10 & \roundnum{0.8843} & \roundnum{0.0958} & \roundnum{0.0996} & \roundnum{0.1192} & \roundnum{0.3445} & \roundnum{7.2559} & \roundnum{3.3983} & \roundnum{1.0067} & \roundnum{0.0912} \\ \hline
10-7 & \roundnum{0.8842} & \roundnum{0.0965} & \roundnum{0.1045} & \roundnum{0.1236} & \roundnum{0.4017} & \roundnum{7.0323} & \roundnum{4.2331} & \roundnum{1.3941} & \roundnum{0.1350} \\ \hline
7-4 & \roundnum{0.8854} & \roundnum{0.0970} & \roundnum{0.1141} & \roundnum{0.1301} &  \roundnum{0.5110} & \roundnum{6.7220} & \roundnum{5.6304} & \roundnum{1.8768}  & \roundnum{0.2093} \\ \hline
4-1 & \roundnum{0.8909} & \roundnum{0.0958} & \roundnum{0.1510} & \roundnum{0.1453} & \roundnum{0.7504} & \roundnum{6.0115} & \roundnum{7.6432} & \roundnum{3.4469}  & \roundnum{0.5503} \\ \hline

    \multicolumn{10}{|c|}{\textit{MagFace}} \\  \hline 
    20+ &\roundnum{ 0.8989} &\roundnum{0.0875} &\roundnum{ 0.0777} &\roundnum{ 0.0929} &\roundnum{ 0.0717} &\roundnum{ 9.0978} &\roundnum{ 0.5283} &\roundnum{0.0373} &\roundnum{0.0000} \\ \hline
16-13 &\roundnum{0.9002} &\roundnum{0.0860} &\roundnum{ 0.1238} &\roundnum{ 0.1168} &\roundnum{ 0.2775} &\roundnum{ 7.5696} &\roundnum{ 2.4918} &\roundnum{0.7057} &\roundnum{0.0675} \\ \hline
13-10 &\roundnum{ 0.9002} &\roundnum{0.0862} &\roundnum{ 0.1293} &\roundnum{ 0.1201} &\roundnum{ 0.3277} &\roundnum{ 7.3741} &\roundnum{ 2.6957} &\roundnum{0.8952} &\roundnum{0.0946} \\ \hline
10-7 &\roundnum{ 0.9003} &\roundnum{0.0862} &\roundnum{ 0.1369} &\roundnum{0.1238} &\roundnum{ 0.3728} &\roundnum{ 7.1598} &\roundnum{ 3.1225} &\roundnum{0.9958} &\roundnum{0.1384} \\ \hline
7-4 &\roundnum{ 0.9017} &\roundnum{0.0857} &\roundnum{ 0.1513} &\roundnum{0.1297} &\roundnum{ 0.4219} &\roundnum{ 6.8269} &\roundnum{ 3.7603} &\roundnum{1.2996} &\roundnum{0.2127} \\ \hline
4-1 &\roundnum{ 0.9085} &\roundnum{0.0825} &\roundnum{ 0.1961} &\roundnum{0.1456} &\roundnum{ 0.6009} &\roundnum{ 6.0202} &\roundnum{ 6.4312} &\roundnum{2.4915} &\roundnum{0.3511} \\ \hline

    \multicolumn{10}{|c|}{\textit{COTS}} \\  \hline 
    20+ &\roundnum{ 0.9843} &\roundnum{0.0209} &\roundnum{ 0.0969} &\roundnum{ 0.1176} &\roundnum{ 0.0909} &\roundnum{ 10.5083} &\roundnum{ 0.9089} &\roundnum{0.0808} &\roundnum{0.0034} \\ \hline
16-13 &\roundnum{ 0.9855} &\roundnum{0.0184} &\roundnum{ 0.2123} &\roundnum{ 0.1895} &\roundnum{ 0.2624} &\roundnum{ 5.7423} &\roundnum{ 2.5433} &\roundnum{0.6493} &\roundnum{0.0505} \\ \hline
13-10 &\roundnum{ 0.9854} &\roundnum{0.0185} &\roundnum{ 0.2293} &\roundnum{0.1969} &\roundnum{ 0.3104} &\roundnum{ 5.4069} &\roundnum{ 2.9905} &\roundnum{0.8343} &\roundnum{0.0639} \\ \hline
10-7 &\roundnum{ 0.9853} &\roundnum{0.0186} &\roundnum{ 0.2448} &\roundnum{0.2039} &\roundnum{ 0.3534} &\roundnum{ 5.1146} &\roundnum{ 3.4616} &\roundnum{1.0361} &\roundnum{0.0942} \\ \hline
7-4 &\roundnum{ 0.9850} &\roundnum{0.0195} &\roundnum{ 0.2577} &\roundnum{0.2105} &\roundnum{ 0.4137} &\roundnum{ 4.8665} &\roundnum{ 4.3155} &\roundnum{1.3959} &\roundnum{0.1715} \\ \hline
4-1 &\roundnum{ 0.9849} &\roundnum{0.0203} &\roundnum{ 0.2688} &\roundnum{0.2197} &\roundnum{ 0.4843} &\roundnum{ 4.5904} &\roundnum{ 5.0283} &\roundnum{1.7826} &\roundnum{0.2253} \\ \hline

    \end{tabular}
    \end{adjustbox}
\end{table}

\subsubsection{Demographic Differentials}

For the analysis of demographics, \ie gender and  race, only results for MagFace will be shown. This is due to the fact that all three different systems show similar patterns, albeit the actual numbers may differ slightly. 

\begin{figure*}[t]
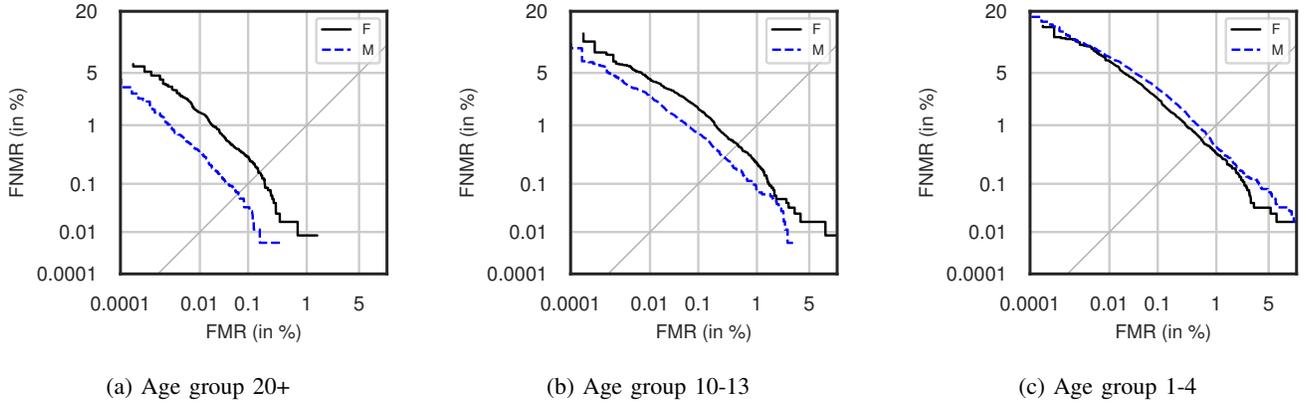

    \centering
    \begin{subfigure}[t]{.33\textwidth}
        \centering
        \scalebox{0.9}{\input{figures/Gender_DET_curve_magface_20.pgf}}
        \caption{Age group 20+}
        \label{fig:gender:det-20+}
    \end{subfigure}%
    \begin{subfigure}[t]{.33\textwidth}
     \centering
         \scalebox{0.9}{\input{figures/Gender_DET_curve_magface_13-10.pgf}}
        \caption{Age group 10-13}
        \label{fig:gender:det-13-10}
    \end{subfigure}
    \begin{subfigure}[t]{.33\textwidth}
        \centering
         \scalebox{0.9}{\input{figures/Gender_DET_curve_magface_4-1.pgf}}
        \caption{Age group 1-4}
        \label{fig:gender:det-4-1}
    \end{subfigure}
    \caption{DET-curves for the Gender datasets at the \textit{1-4}, \textit{10-13} \& \textit{20+} age groups using \textit{MagFace}. 
    }
    \label{fig:gender-det}\vspace{-0.3cm}
\end{figure*}

In table~\ref{tab:gender:det-statistics}, obtained results for the gender subset are summarized. it can be observed that the \texttt{$d'$} values are larger for males than females when looking at the age groups 20+, 16-13 and 13-10; thus, the systems are better at distinguishing mated and non-mated samples. The value of the younger age groups is a bit larger for females. The non-mated mean values are generally larger for males across all age groups, but for mated, the values are very similar for both males and females.  The DET-curves for the three age groups 20+, 10-13 and 1-4 divided into gender are depicted in figure~\ref{fig:gender-det}. For the first five age groups, the EER is lower for males than females, but for the last age group, ages 4-1, the opposite can be observed. This is the same phenomenon observed from the DET curves in figure~\ref{fig:gender-det}, where the male subset performed better than the female subset in all but the youngest age group.

\begin{figure*}[t]
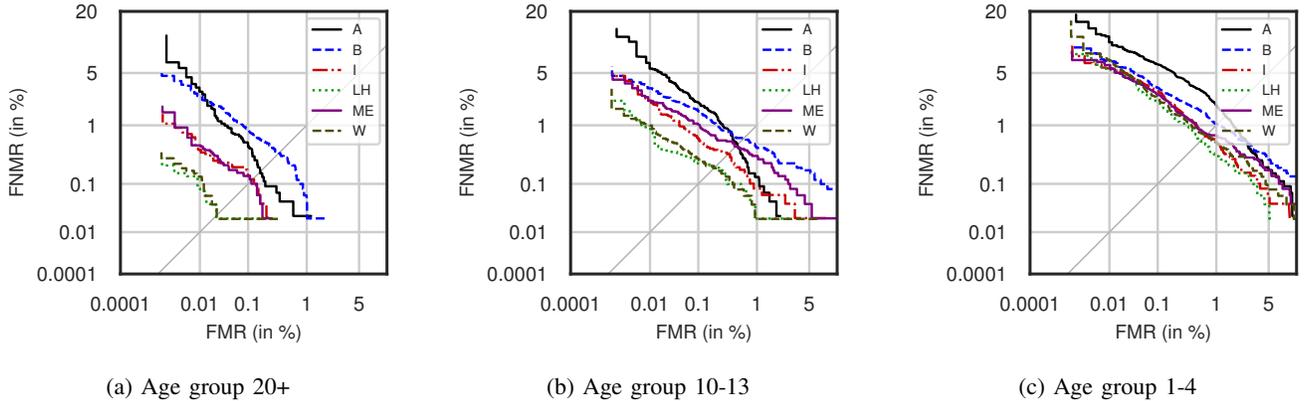

    \centering
    \begin{subfigure}[t]{.33\textwidth}
        \centering
        \scalebox{0.9}{\input{figures/Race_DET_curve_magface_20.pgf}}
        \caption{Age group 20+}
        \label{fig:race:det-20+}
    \end{subfigure}%
    \begin{subfigure}[t]{.33\textwidth}
     \centering
         \scalebox{0.9}{\input{figures/Race_DET_curve_magface_13-10.pgf}}
        \caption{Age group 10-13}
        \label{fig:race:det-13-10}
    \end{subfigure}
    \begin{subfigure}[t]{.33\textwidth}
        \centering
         \scalebox{0.9}{\input{figures/Race_DET_curve_magface_4-1.pgf}}
        \caption{Age group 1-4}
        \label{fig:race:det-4-1}
    \end{subfigure}
    \caption{DET-curves for the Race datasets at the \textit{1-4}, \textit{10-13} \& \textit{20+} age groups using \textit{MagFace}. 
    }
    \label{fig:det-race}
\end{figure*}

Regarding race, obtained results are summarised in table~\ref{tab:race-det}. It should be noted that only the non-mated distribution statistics are shown here, as the mated statistics are very similar and would produce an overwhelmingly large table. The table contains several interesting observations, such as the \texttt{$d'$} score being largest for subjects of white race in all age groups, except the adult one where Latino-Hispanic has a slightly larger one. Subjects of the black race always have the smallest \texttt{$d'$} score, closely followed by Indians. The mean, standard deviation, and median change significantly depending on the race and age. It can be observed that the ones with the Asian race have the most significant standard deviation for all but the oldest age group. The corresponding DET-curves can be seen in Figure~\ref{fig:det-race}. The performance worsens for all the races in the youngest age groups but still with a similar hierarchy of the worst to best performing races.

\setlength{\tabcolsep}{5pt}
\begin{table}[!htbp]
    \caption{Different biometric performance metrics with \textit{MagFace} from the gender-divided dataset.}
    \label{tab:gender:det-statistics}
    \begin{adjustbox}{max width=1\columnwidth,center}
 \begin{tabular}{|c|c|c|c|c|c|c|c|c|c|c|}
        \hline
        \textbf{Age}  & \multirow{2}{*}{\textbf{Gender}} &  \multicolumn{2}{c|}{\textbf{Mated}}  & \multicolumn{2}{c|}{\textbf{Non-mated}}  & \textit{EER} & \multirow{2}{*}{\textit{d'}} & \multicolumn{3}{c|}{\textit{FNMR at FMR} (\%)}    \\ %
        \textbf{groups}& &  $\mu$ & $\sigma$ &  $\mu$ & $\sigma$ & (\%) & & \textit{0.01} & \textit{0.1}  & \textit{ 1} \\ \hline
        \multirow{2}{*}{20+} & F & \roundnum{0.8971} & \roundnum{0.0892} & \roundnum{0.0853} & \roundnum{0.1058}  & \roundnum{0.1590} &  \roundnum{8.2953} & \roundnum{1.5403} & \roundnum{0.3097} & \roundnum{0.0084} \\ \cline{2-11}
 & M  &  \roundnum{0.9001} & \roundnum{0.0864} &  \roundnum{0.0869} & \roundnum{0.0956} & \roundnum{0.0637} & \roundnum{8.9237} & \roundnum{0.3640} & \roundnum{0.0341} & \roundnum{0.0000} \\
\hline\multirow{2}{*}{16-13} & F & \roundnum{0.8994} & \roundnum{0.0902} & \roundnum{0.1195} & \roundnum{0.1228} & \roundnum{0.4350}  & \roundnum{7.2398} & \roundnum{4.1412 } & \roundnum{1.5645} & \roundnum{0.1924} \\ \cline{2-11}
 & M & \roundnum{0.9008} & \roundnum{0.0830} & \roundnum{0.1516} & \roundnum{0.1182} & \roundnum{0.2494} & \roundnum{7.3331} & \roundnum{2.0154} & \roundnum{0.6058} & \roundnum{0.0566} \\
\hline\multirow{2}{*}{13-10} & F & \roundnum{0.9001} & \roundnum{0.0900} & \roundnum{0.1251} & \roundnum{0.1257} & \roundnum{0.4957} & \roundnum{7.0916}  & \roundnum{4.2368} & \roundnum{1.7332} & \roundnum{0.2428} \\ \cline{2-11}
 & M & \roundnum{0.9003} & \roundnum{0.0835} & \roundnum{0.1565} & \roundnum{0.1213} & \roundnum{0.3001} & \roundnum{7.1427} & \roundnum{2.5198} & \roundnum{0.7475} & \roundnum{0.0680} \\
\hline\multirow{2}{*}{10-7} & F & \roundnum{0.9007} & \roundnum{0.0895} & \roundnum{0.1317} & \roundnum{0.1282} & \roundnum{0.5187} & \roundnum{6.9551} & \roundnum{4.8033} & \roundnum{1.8661} & \roundnum{0.3013} \\ \cline{2-11}
 & M & \roundnum{0.9001} & \roundnum{0.0838} &  \roundnum{0.1640} & \roundnum{0.1250} & \roundnum{0.3508} & \roundnum{6.9165} & \roundnum{3.0836} & \roundnum{0.8940} & \roundnum{0.1301} \\
\hline\multirow{2}{*}{7-4} & F & \roundnum{0.9017} & \roundnum{0.0887} &  \roundnum{0.1416} & \roundnum{0.1313} & \roundnum{0.5604} & \roundnum{6.7849} & \roundnum{5.1857} & \roundnum{2.0659} & \roundnum{0.3262} \\ \cline{2-11}
 & M & \roundnum{0.9016} & \roundnum{0.0836} & \roundnum{0.1824} & \roundnum{0.1318} & \roundnum{0.4377} & \roundnum{6.5163} & \roundnum{4.0014} & \roundnum{1.2734} & \roundnum{0.1924} \\
\hline\multirow{2}{*}{4-1} & F & \roundnum{0.9056} & \roundnum{0.0856} & \roundnum{0.1692} & \roundnum{0.13800} & \roundnum{0.5858} & \roundnum{6.4128} & \roundnum{6.7754} & \roundnum{2.3923} & \roundnum{0.3513} \\ \cline{2-11}
 & M & \roundnum{0.9104} & \roundnum{0.0803} &  \roundnum{0.2444} & \roundnum{0.1481} & \roundnum{0.7081} & \roundnum{5.5924} & \roundnum{7.4833} & \roundnum{3.2152} & \roundnum{0.4472} \\
\hline

    \end{tabular}
    \end{adjustbox}
\end{table}

\setlength{\tabcolsep}{7pt}
\begin{table}[!htbp]
\caption{Different biometric measures from using \textit{MagFace} on the race-divided dataset.}
    \label{tab:race-det}
    \begin{adjustbox}{max width=1\columnwidth,center}
 \begin{tabular}{|c|c|c|c|c|c|c|c|c|}
        \hline
        \textbf{Age}  & \multirow{2}{*}{\textbf{Race}} & \multicolumn{2}{c|}{\textbf{Non-mated}}  & \textit{EER} & \multirow{2}{*}{\textit{d'}} & \multicolumn{3}{c|}{\textit{FNMR at FMR of (\%)}}    \\ %
        \textbf{groups} & &  $\mu$ & $\sigma$ & (\%) & & \textit{0.01} & \textit{0.1}  & \textit{1} \\ \hline
        \multirow{6}{*}{20+} & A  & \roundnum{0.1101} & \roundnum{0.1072} & \roundnum{0.1627} & \roundnum{8.1485} & \roundnum{2.9438} & \roundnum{0.4719} & \roundnum{0.0225} \\ \cline{2-9}
 & B  & \roundnum{0.1862} & \roundnum{0.1307} & \roundnum{0.3586} & \roundnum{6.4519}  & \roundnum{2.5922} & \roundnum{0.8574} & \roundnum{0.0399} \\ \cline{2-9}
 & I  & \roundnum{0.2188} & \roundnum{0.1158} & \roundnum{0.1217} & \roundnum{7.2459}  & \roundnum{0.3664} & \roundnum{0.1425} & \roundnum{0.0000} \\ \cline{2-9}
 & LH  & \roundnum{0.1136} & \roundnum{0.0946} & \roundnum{0.0214} & \roundnum{8.4609}  & \roundnum{0.0794} & \roundnum{0.0198} & \roundnum{0.0000} \\ \cline{2-9}
 & ME  & \roundnum{0.1182} & \roundnum{0.1065} & \roundnum{0.1080} & \roundnum{7.8258}  & \roundnum{0.4620} & \roundnum{0.1406} & \roundnum{0.0000} \\ \cline{2-9}
 & W  & \roundnum{0.1056} & \roundnum{0.0906} & \roundnum{0.0214} & \roundnum{8.3885}  & \roundnum{0.1364} & \roundnum{0.0195} & \roundnum{0.0000} \\
\hline\multirow{6}{*}{16-13} & A  & \roundnum{0.1488} & \roundnum{0.1339} & \roundnum{0.3853} & \roundnum{6.7241}  & \roundnum{5.2974} & \roundnum{1.9978} & \roundnum{0.0673} \\ \cline{2-9}
 & B  & \roundnum{0.2963} & \roundnum{0.1314} & \roundnum{0.5553} & \roundnum{5.5653}  & \roundnum{3.5310} & \roundnum{1.3291} & \roundnum{0.3571} \\ \cline{2-9}
 & I  & \roundnum{0.3616} & \roundnum{0.1216} & \roundnum{0.2844} & \roundnum{5.7042} &  \roundnum{1.4780} & \roundnum{0.4657} & \roundnum{0.0810} \\ \cline{2-9}
 & LH  & \roundnum{0.1771} & \roundnum{0.1108} & \roundnum{0.1976} & \roundnum{7.1590} & \roundnum{0.8119} & \roundnum{0.2376} & \roundnum{0.0198} \\ \cline{2-9}
 & ME  & \roundnum{0.1977} & \roundnum{0.1240} & \roundnum{0.4048} & \roundnum{6.4267} & \roundnum{1.9419} & \roundnum{0.8208} & \roundnum{0.2402} \\ \cline{2-9}
 & W  & \roundnum{0.1434} &\roundnum{ 0.1068} & \roundnum{0.1406} & \roundnum{7.2574} & \roundnum{0.7592} & \roundnum{0.2141} & \roundnum{0.0195} \\
\hline\multirow{6}{*}{13-10} & A  & \roundnum{0.1583} & \roundnum{0.1388} & \roundnum{0.4826} & \roundnum{6.4869} & \roundnum{5.6142} & \roundnum{2.1109} & \roundnum{0.1347} \\ \cline{2-9}
 & B  & \roundnum{0.3075} & \roundnum{0.1312} & \roundnum{0.6355} & \roundnum{5.4346} & \roundnum{3.2572} & \roundnum{1.6286} & \roundnum{0.4369} \\ \cline{2-9}
 & I  & \roundnum{0.3754} & \roundnum{0.1210} & \roundnum{0.3440} & \roundnum{5.5983} & \roundnum{2.0876} & \roundnum{0.6080} & \roundnum{0.0811} \\ \cline{2-9}
 & LH  & \roundnum{0.1851} & \roundnum{0.1128} & \roundnum{0.1976} & \roundnum{7.0189} & \roundnum{0.7723} & \roundnum{0.2376} & \roundnum{0.0198} \\ \cline{2-9}
 & ME  & \roundnum{0.2038} & \roundnum{0.1264} & \roundnum{0.4601} & \roundnum{6.2794} & \roundnum{2.2218} & \roundnum{1.0408} & \roundnum{0.3203} \\ \cline{2-9}
 & W  & \roundnum{0.1491} & \roundnum{0.1087} & \roundnum{0.1950} & \roundnum{7.1094} & \roundnum{0.9932} & \roundnum{0.2921} & \roundnum{0.0195} \\
\hline\multirow{6}{*}{10-7} & A  & \roundnum{0.1711} & \roundnum{0.1437} & \roundnum{0.6059} & \roundnum{6.2255} & \roundnum{7.8392} & \roundnum{2.4034} & \roundnum{0.2471} \\ \cline{2-9}
 & B  & \roundnum{0.3168} & \roundnum{0.1312} & \roundnum{0.6137} & \roundnum{5.3329} & \roundnum{3.5665} & \roundnum{1.8229} & \roundnum{0.5350} \\ \cline{2-9}
 & I  & \roundnum{0.3874} & \roundnum{0.1205} & \roundnum{0.4123} & \roundnum{5.4950} & \roundnum{2.2667} & \roundnum{0.8298} & \roundnum{0.0810} \\ \cline{2-9}
 & LH  & \roundnum{0.1964} & \roundnum{0.1153} & \roundnum{0.2196} & \roundnum{6.8434} & \roundnum{0.5939} & \roundnum{0.2574} & \roundnum{0.0396} \\ \cline{2-9}
 & ME  & \roundnum{0.2112} & \roundnum{0.1286} & \roundnum{0.5399} & \roundnum{6.1425} & \roundnum{2.8028} & \roundnum{1.1411} &\roundnum{0.3804} \\ \cline{2-9}
 & W  & \roundnum{0.1564} & \roundnum{0.1110} & \roundnum{0.2389} & \roundnum{6.9499} & \roundnum{1.2262} & \roundnum{0.3893} & \roundnum{0.0389} \\
\hline\multirow{6}{*}{7-4} & A  & \roundnum{0.1939} & \roundnum{0.1534} & \roundnum{0.8528} & \roundnum{5.7786} & \roundnum{8.7826} & \roundnum{3.7736} & \roundnum{0.6739} \\ \cline{2-9}
 & B  & \roundnum{0.3285} & \roundnum{0.1327} & \roundnum{0.7708} & \roundnum{5.1749} & \roundnum{4.7261} & \roundnum{2.2543} & \roundnum{0.6723} \\ \cline{2-9}
 & I  & \roundnum{0.3951} & \roundnum{0.1216} & \roundnum{0.4046} & \roundnum{5.4071} & \roundnum{2.5719} & \roundnum{1.0531} & \roundnum{0.2025} \\ \cline{2-9}
 & LH  & \roundnum{0.2135} & \roundnum{0.1197} & \roundnum{0.2422} & \roundnum{6.5680} & \roundnum{1.2874} & \roundnum{0.3961} & \roundnum{0.0792} \\ \cline{2-9}
 & ME  & \roundnum{0.2264} & \roundnum{0.1324} & \roundnum{0.6201} & \roundnum{5.9140} & \roundnum{4.0232} & \roundnum{1.5212} & \roundnum{0.4604} \\ \cline{2-9}
 & W  & \roundnum{0.1738} & \roundnum{0.1159} & \roundnum{0.3212} & \roundnum{6.6291} & \roundnum{2.3768} & \roundnum{0.6429} & \roundnum{0.1364} \\
\hline\multirow{6}{*}{4-1} & A  & \roundnum{0.2498} & \roundnum{0.1747} & \roundnum{1.3702} & \roundnum{4.9065} & \roundnum{10.6694} & \roundnum{6.3118} & \roundnum{1.9317} \\ \cline{2-9}
 & B  & \roundnum{0.3597} & \roundnum{0.1387} & \roundnum{1.0460} & \roundnum{4.8028} & \roundnum{6.7930} & \roundnum{3.3175} & \roundnum{1.0664} \\ \cline{2-9}
 & I  & \roundnum{0.3998} & \roundnum{0.1334} & \roundnum{0.6763} & \roundnum{5.0348} & \roundnum{5.6524} & \roundnum{3.0186} & \roundnum{0.5875} \\ \cline{2-9}
 & LH  & \roundnum{0.2539} & \roundnum{0.1393} & \roundnum{0.5748} & \roundnum{5.6898} & \roundnum{5.4124} & \roundnum{2.1610} & \roundnum{0.3172} \\ \cline{2-9}
 & ME  & \roundnum{0.2720} & \roundnum{0.1429} & \roundnum{0.7209} & \roundnum{5.3645} & \roundnum{5.5912} & \roundnum{2.6854} & \roundnum{0.6814} \\ \cline{2-9}
 & W  & \roundnum{0.2198} & \roundnum{0.1354} &  \roundnum{0.6810} & \roundnum{5.7845} & \roundnum{6.8939} & \roundnum{2.4732} & \roundnum{0.5453} \\
\hline
    \end{tabular}
    \end{adjustbox}
\end{table}

\section{Discussion}\label{sec:discussion}

From the observed results of the full dataset, the mated scores are stable across the different age groups and in general have quite a high mean. The progression highly impacts the non-mated scores across age groups, which causes a performance decrease as the subjects gets younger when verification metrics are measured. This drop in performance was a common pattern across all three tested face recognition systems. Performance significantly decreases as the subjects get younger, with a notable increase in EER. A common threshold in biometric verification is having a FMR of 0.1\%~\cite{FRONTEX2015} and looking at MagFace in table~\ref{tab:full-dataset-det-stats}, this results in a practical FNMR of 0.04\% for the adults. But by progressing these same adults down to an age group of the youngest children of age 1-4, it rises to 2.49\%. This change in score demonstrates how large a performance decrease can be observed when the same identities are younger. On the youngest children, the COTS system showed the best performance, according to EER values. Although when looking at the adults COTS performed the worst. It is unknown what kind of data is used in training this system, but it could indicate that the system has seen images of young children before.
As this analysis of the performance across ages is based on synthetic data, the question arises whether these same observed results would happen if it was tested on facial images of persons at different ages. 
As mentioned in section~\ref{sec:related}, several studies on child face recognition were described in which a performance drop was also observed in real data of children, compared to the performance of adults. Notably, in the recent paper~\cite{YLFW}, a performance decrease in younger children is seen compared to adults. The authors also note that children are harder to discriminate for the different facial recognition systems that they test. They do see a performance increase by fine-tuning a system on their child database. 
These results are comparable with the results observed in this work, but this dataset has the benefit of being synthetic.  

It was also observed how subjects of Black and Asian race in general perform worse than ones of White and Latino-Hispanic. It was further seen that all races have a performance decrease as they get younger. In~\cite{NISTDemographicEffects}, Grother~\etal have performed a vendor test with a specific focus on the performance and bias of commercial face recognition systems concerning demographics. In the report, they note several of the same observations regarding race and age. Similar findings were made in~\cite{RFW}. Overall the results indicate that facial recognition systems are not robust to younger subjects and that racial and gender bias is a general problem across age groups.

\begin{figure}[!htbp]
    \centering
    \includegraphics[width=0.9\linewidth]{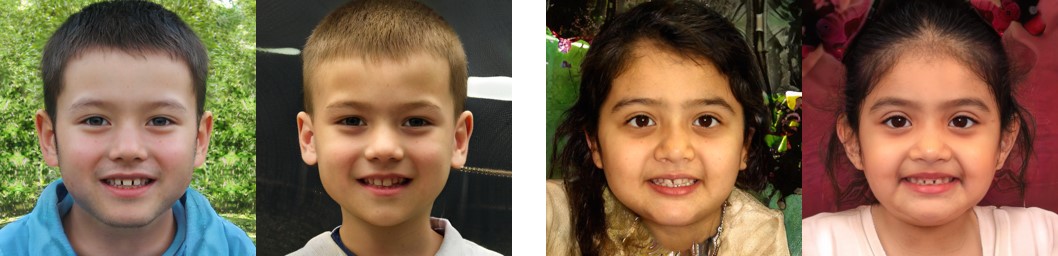}
    \caption{Example pairs of synthetic children subjects (ages 7-10) with a high non-mated score according to \textit{MagFace}.}
    \label{fig:7-10:high-non-mated}
\end{figure}

\begin{figure}[!htbp]
    \centering
    \includegraphics[width=0.9\linewidth]{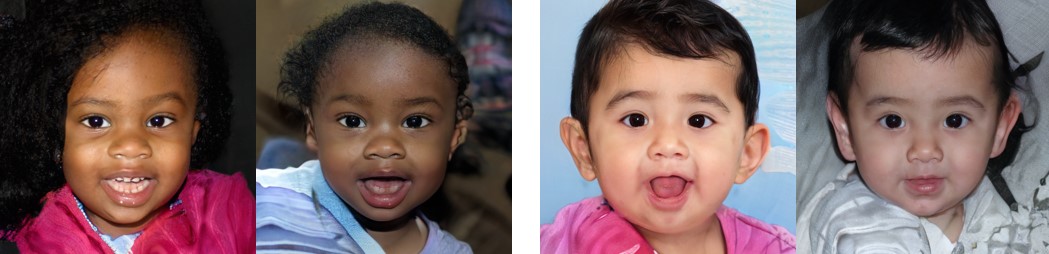}
    \caption{Example pairs of synthetic children subjects (ages 1-4) with a high non-mated score according to \textit{MagFace}.}
    \label{fig:1-4:high-non-mated}
\end{figure}

In figure~\ref{fig:7-10:high-non-mated}, pairs of subjects from ages 7-10 with a high non-mated score can be seen. As seen, the pairs of subjects has the same gender and race. Similarly, pairs of subjects from the youngest age group (ages 1-4) with a high non-mated score can be seen in figure~\ref{fig:1-4:high-non-mated}. In this particular age group, false matches across gender and race have also been observed.

\begin{figure}[!htbp]
    \centering
    \includegraphics[width=0.9\linewidth]{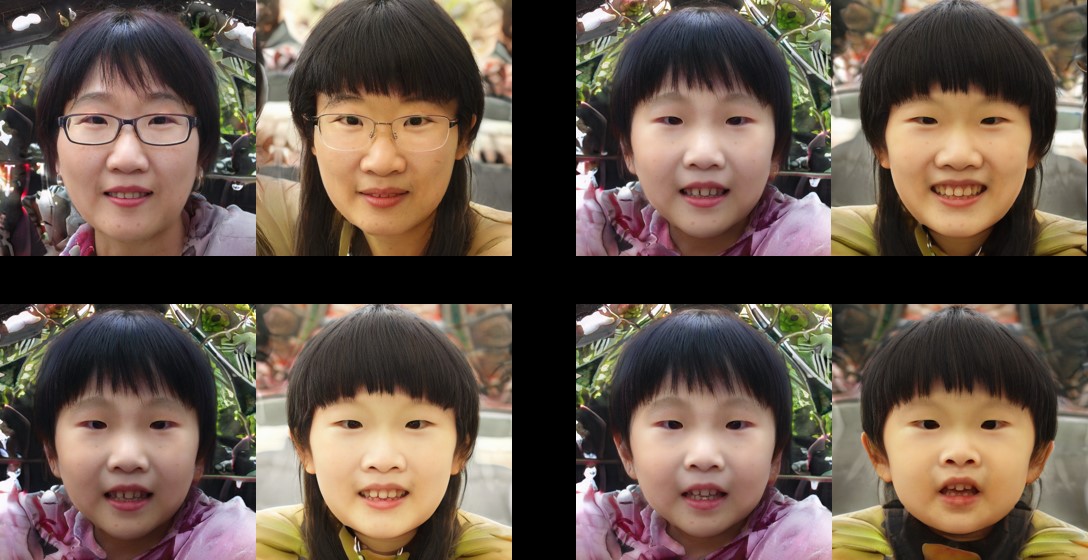}
    \caption{An example of two subjects who have a high non-mated score throughout all age groups, according to \textit{MagFace}.}
    \label{fig:results-phenomena}
\end{figure}

\begin{figure}
    \centering
    \includegraphics[width=\linewidth]{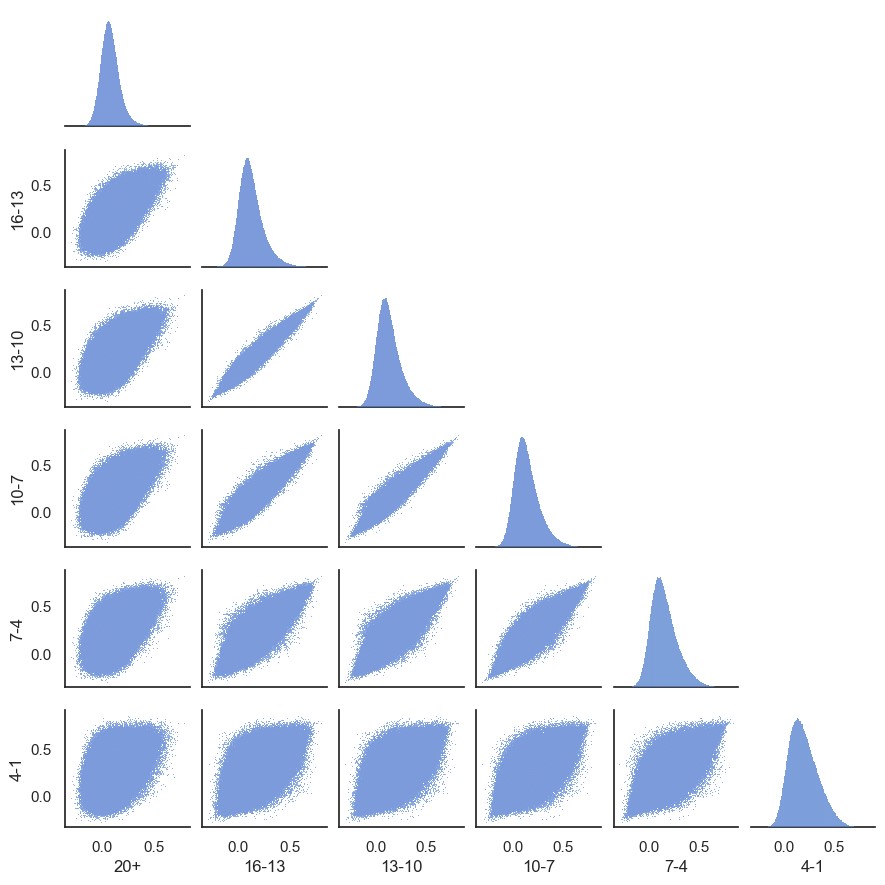}
    \caption{Pair-wise scatter plots between the different age groups showing the relationship between the non-mated comparison score with \textit{MagFace}.}
    \label{fig:scatter-matrix}
\end{figure}

An example where the same two subjects have a high non-mated score in all of the different age groups can be seen in figure~\ref{fig:results-phenomena}. Here the top left image is from the adult age group while the bottom right is from the youngest child group. To investigate this phenomena further the scatter matrix in Figure~\ref{fig:scatter-matrix} was made. 

In the figure, each of the non-mated scores of one age group was plotted against the same non-mated comparison of all other age groups. So if some subject \texttt{s1} is compared with some other subject \texttt{s2} in age group 20+ and produces some score, then the same two subjects have comparison scores in all of the other age groups. All of these scores can then be plotted pairwise to see if there is a correlation between scores of the same subjects across age groups. An example can be seen by looking at the $x$-axis at ages 20+ and the $y$-axis at ages 16-13, where there is a positive correlation. An even stronger correlation can be observed at $x$-axis 16-13 and $y$-axis 13-10, which may be because the age groups are much closer in age than those of 20+ and 16-13. This positive correlation continues down the whole diagonal, which indeed tells us that there is a tendency for the same pairs of non-mated comparison scores being correlated across age groups. 

\section{Conclusions}\label{sec:conclusion}
In this work we introduced the HDA-SynChildFaces database, a synthetic database of demographically balanced face images of children across various age groups including common intra-class variations. 

From experiments conducted on  HDA-SynChildFaces the following  key findings were obtained: 
\begin{itemize}
    \item The mated scores are on average not impacted much by face age progression in all tested face recognition systems.
    \item The non-mated scores on average become significantly higher proportional to the age in all tested systems.
    \item The EER and different FNMR rates at relevant FMR rates increase proportional to the age in all tested systems
    \item Subjects classified as female have higher EER as well as FNMR rates than males at all age groups, with some exceptions in the youngest group (ages 1-4).
    \item The race of the subjects has an impact on the systems, and the performance across all races gets worse as the age of the subjects decrease. Especially subjects of Black and Asian have high EER and FNMR rates compared to subjects of White and Latino-Hispanic, also as children.
\end{itemize}

\section*{Acknowledgement}
This research is based upon work supported by the H2020 TReSPAsS-ETN Marie Sk\l{}odowska-Curie early training network (grant agreement 860813), by the Hessian Ministry of the Interior and Sport in the course of the Bio4ensics project and by the German Federal Ministry of Education and Research and the Hessian Ministry of Higher Education, Research, Science and the Arts within their joint support of the National Research Center for Applied Cybersecurity ATHENE.

\bibliographystyle{IEEEtran.bst}
\bibliography{bibli}
\end{document}